%% file: main.tex
\documentclass{article}

\usepackage[preprint]{neurips_2022}

\usepackage[utf8]{inputenc}
\usepackage[T1]{fontenc}

\usepackage{newtxtext,newtxmath}

\usepackage[dvipsnames]{xcolor}   
\definecolor{linkColor}{rgb}{0.2,0.4,0.6} 
\usepackage{graphicx}
\usepackage{booktabs}
\usepackage{amsmath}
\usepackage{mathtools}
\usepackage{nicefrac}
\usepackage{multirow}
\usepackage{wrapfig}
\usepackage{float}
\usepackage{bbm}
\usepackage{enumitem}
\usepackage{url}
\usepackage{pifont}
\usepackage{wasysym}
\usepackage{arydshln}
\usepackage{listings}
\usepackage[most]{tcolorbox}      
\usepackage{adjustbox}
\usepackage{setspace}
\usepackage[normalem]{ulem}       
\usepackage{capt-of}
\usepackage{algorithm}
\usepackage{algpseudocode}        
\usepackage{mdframed}
\usepackage{subcaption}           
\usepackage{makecell}
\usepackage{colortbl}
\usepackage{hyperref}

\hypersetup{
  colorlinks=true,
  linkcolor=linkColor,
  citecolor=linkColor,
  urlcolor=linkColor
}

\definecolor{deemph}{rgb}{0.4,0.4,0.4}        

\newcommand\ours{{\texttt{RSPO}}}
\newcommand\oursfull{{\texttt{Router-Shift Policy Optimization}}}
\definecolor{darkblue}{rgb}{0, 0, 0.6}

\lstdefinestyle{mypython}{
  language=Python,
  basicstyle=\ttfamily\footnotesize,
  keywordstyle=\bfseries,
  commentstyle=\itshape,
  stringstyle=,
  columns=fullflexible,
  keepspaces=true,
  showstringspaces=false,
  tabsize=4,
  breaklines=true,
}

\title{Towards Stable and Effective Reinforcement\\ Learning for Mixture-of-Experts}

\author{%
Di Zhang\textsuperscript{\dag\ddag}\quad
Xun Wu\textsuperscript{\dag}\quad
Shaohan Huang\textsuperscript{\dag}\quad
Lingjie Jiang\textsuperscript{\dag\ddag}\quad
Yaru Hao\textsuperscript{\dag}\quad\\
\bfseries
Li Dong\textsuperscript{\dag}
Zewen Chi\textsuperscript{\dag}\quad
Zhifang Sui\textsuperscript{\ddag}\quad
Furu Wei\textsuperscript{\dag}\\[0.4em]
\textsuperscript{\dag}\,Microsoft Research \quad
\textsuperscript{\ddag}\,Peking University\\
\href{https://aka.ms/GeneralAI}{https://aka.ms/GeneralAI}
}

\begin{document}
\maketitle

\begin{abstract}
Reinforcement learning with verifiable rewards (RLVR) has emerged as a powerful paradigm for improving reasoning capabilities. However, training RLVR with Mixture-of-Experts (MoE) policies remains fragile and is often prone to reward collapse.
We identify a MoE-specific source of instability, referred to as router shift (RS), where changes in expert routing across policy updates exacerbate off-policy mismatch. This effect leads to increasingly volatile importance-ratio signals and bursty clipping behavior, which consistently precede training collapse.
Motivated by this diagnosis, we propose Router-Shift Policy Optimization (RSPO). RSPO computes a per-token router-shift ratio conditioned on the previously activated experts, applies stop-gradient and a lower-bound floor, and softly rescales importance ratios prior to clipping and aggregation. This design explicitly accounts for routing-induced distributional drift during off-policy optimization.
%
We evaluate the effect of RSPO under two settings: a synthetic countdown task and real-world reasoning tasks on MATH and Code. Across both settings, RSPO achieves better performance and exhibits greater stability compared to recent MoE-based RLVR methods.
%
\end{abstract}

\begin{figure*}[!h]
  \centering
  \includegraphics[width=0.82\linewidth]{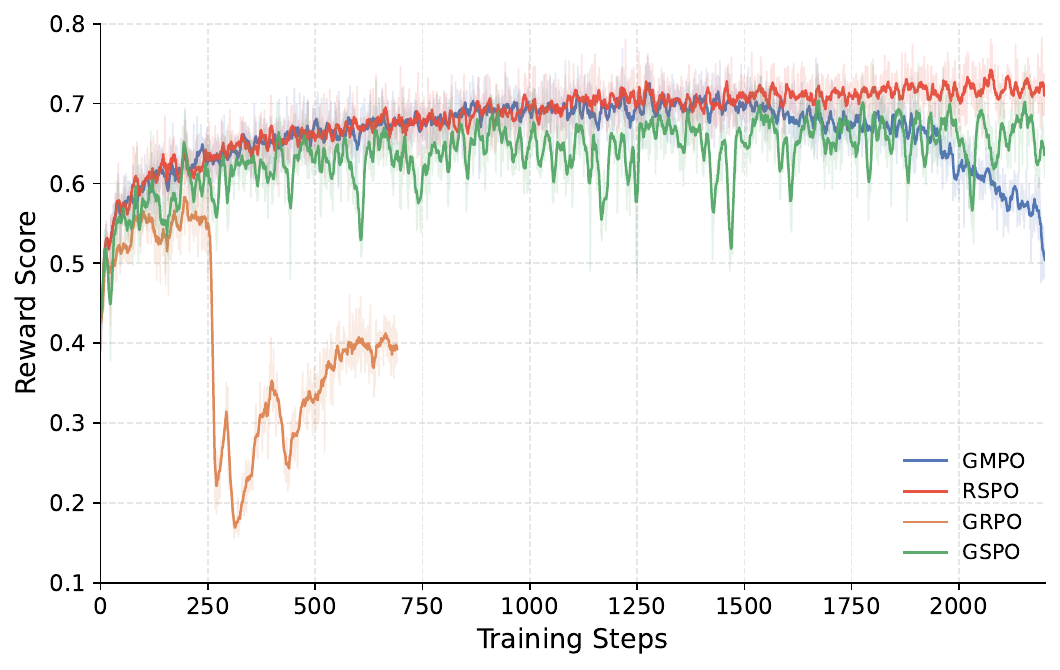}
  \caption{
  \textbf{Training instability on MoE.}
  Training reward versus step for GRPO and GRPO-style stabilizations (GSPO/GMPO/RSPO) on Qwen2.5-MoE under the Countdown RLVR setting. \textbf{Our RSPO achieves better performance while exhibiting stronger stability.}}
  \label{fig:diagnosis_collapse}
  \vspace{-2mm}
\end{figure*}

\input{sections_new/intro}
\input{sections_new/prelimi}
\input{sections_new/instability}

\input{sections_new/method}
\input{sections_new/exp_new}

\input{sections/related}

\input{sections_new/conclu}

\appendix
\input{sections_new/append}
\label{sec:appendix}

\bibliographystyle{alpha} 
\bibliography{main}

\end{document}

%% file: sections_new/intro.tex
\section{Introduction}

Reinforcement learning with verifiable rewards (RLVR) has become a central approach for post-training large language models (LLMs) in reasoning and code generation. By relying on deterministic, rule-based verifiers that provide sparse correctness signals, RLVR has been shown to elicit strong reasoning behaviors and achieve substantial gains on challenging tasks such as mathematical problem solving and program synthesis \citep{o1,guo2025deepseek,yang2025qwen3,team2025kimi,chen2025minimax}.
In parallel, Mixture-of-Experts (MoE) architectures offer an efficient scaling mechanism by activating only a small subset of experts per token \citep{fedus2022switch}, making them particularly attractive for large-scale RLVR training where computational efficiency is critical.

Despite these advances, directly applying RLVR to MoE models remains brittle and
often exhibits severe training instability \citep{zheng2025group,chen2025minimax,yang2025qwen3}.
As illustrated in Fig.~\ref{fig:diagnosis_collapse}, GRPO can suffer from abrupt reward collapse on MoE models.
A key MoE-specific challenge is \emph{router drift} (also referred to as router fluctuation): the activated experts and
their routing probabilities for the \emph{same} token may change substantially across policy updates
\citep{dai2022stablemoe,zheng2025group}.
Such routing changes can amplify off-policy mismatch and destabilize optimization.
Moreover, RLVR commonly uses sequence-level rewards (binary correctness for an entire solution), while many practical
implementations still apply token-level importance ratios and clipping, leading to additional variance and further
compounding instability.

Existing stabilizations address this problem only partially.
GSPO~\citep{zheng2025group} and GMPO~\citep{zhao2025geometric} reduce variance mismatch by using sequence-level likelihood ratios or geometric-mean aggregation,
which improves robustness to token-level outliers.
However, these methods do not explicitly control the impact of \emph{routing drift} on off-policy updates.
A seemingly straightforward alternative is to constrain routing directly, e.g., freezing the router or replaying routing~\citep{zheng2025group} decisions across updates. In our experiments, these rigid strategies are unsatisfactory: freezing harms router adaptivity to the RL objective,
while replay-based constraints limit router exploration and can degrade performance (see Sec.~\ref{sec:exp_alternatives} and Appendix~\ref{app:alternatives}).

In this work, we provide a concise diagnosis that links routing instability to optimization instability.
Using lightweight training-time signals (without logging full token-level ratio distributions), we show that routing stability degrades over training and coincides with increasingly volatile off-policy mismatch signals and bursty clipping activity, which together increase the risk of reward collapse (Sec.~\ref{sec:diagnosis}).
This diagnosis motivates a targeted intervention: to stabilize MoE off-policy RL, we should directly reduce the
influence of tokens whose routing behavior drifts substantially across updates, \emph{without} hard-freezing the router
or fully replaying routing decisions.

Motivated by this, we propose \oursfull{} (\ours{}), a router-aware modification to GRPO-style objectives.
\ours{} computes a per-token \emph{router-shift ratio} from router scores on the \emph{old activated experts} across MoE layers, applies a simple processing step (stop-gradient and lower-bound flooring), and multiplies the resulting
trust weight into the importance ratio before the usual clipping and aggregation.
This yields a \emph{soft} adjustment mechanism: tokens with severe routing deviations contribute less to the policy
update, mitigating routing-induced off-policy mismatch while preserving router adaptivity.

We evaluate \ours{} under two complementary regimes.
In a small-scale diagnostic setting (Qwen2.5-MoE on Countdown) (See Fig.~\ref{fig:diagnosis_collapse}), router-shift weighting consistently stabilizes GRPO and
its variants when used as a plug-in component.
In a large-scale benchmark setting (Qwen3-30B-A3B), \ours{} (GMPO+RS) improves downstream Pass@1 on both \textbf{math}
and \textbf{code} benchmarks and yields more stable training-time routing/optimization diagnostics compared to GRPO.
Overall, our results highlight the importance of router-aware stabilization for MoE RLVR. Our main contributions are:
\begin{itemize}[leftmargin=1.5em,itemsep=0pt,parsep=0.2em,topsep=0.0em,partopsep=0.0em]
    \item \textbf{Diagnosis of MoE instability in off-policy RLVR.}
    We provide measurable evidence linking router drift to volatile off-policy mismatch signals and bursty clipping
    behavior that precede reward collapse.
    \item \textbf{Router-aware soft stabilization.}
    We propose \ours{}, which computes a per-token router-shift ratio from old activated experts and uses it as a
    detached, floored trust weight to rescale importance ratios prior to clipping/aggregation, preserving router adaptivity.
    \item \textbf{Empirical validation at two scales.}
    We show that router-shift weighting acts as a plug-in stabilization module on Qwen2.5-MoE (Countdown) and that
    \ours{} improves stability and final performance on Qwen3-30B-A3B across both math and code benchmarks.
\end{itemize}

%% file: sections_new/prelimi.tex
\section{Preliminaries}
\label{sec:prelim}

\paragraph{Group Relative Policy Optimization (GRPO).}
Given a query $x$, GRPO samples a group of $G$ responses
$\{y_i\}_{i=1}^{G}\sim \pi_{\theta_{\text{old}}}(\cdot\mid x)$ and computes
group-relative advantages from a scalar reward $r(x,y_i)$.
Let $\hat{A}_i$ denote the normalized group advantage (shared across tokens in $y_i$),
and define the token-level importance ratio
\begin{equation}
w_{i,t}(\theta)\triangleq
\frac{\pi_{\theta}(y_{i,t}\mid x, y_{i,<t})}
{\pi_{\theta_{\text{old}}}(y_{i,t}\mid x, y_{i,<t})}.
\label{eq:grpo_token_ratio}
\end{equation}
GRPO optimizes a PPO-style clipped surrogate at the token level:
\begin{equation}
\resizebox{0.43\textwidth}{!}{%
    $
\ell^{\textsc{grpo}}_{i,t}(\theta)\triangleq
\min\!\Big(
w_{i,t}(\theta)\hat{A}_i,\;
\mathrm{clip}\!\big(w_{i,t}(\theta),\,1-\epsilon,\,1+\epsilon\big)\hat{A}_i
\Big),
$
}
\label{eq:grpo_surrogate}
\end{equation}
and averages it over tokens and group samples (full objective in
Appendix~\ref{app:baseline_objectives}).

\paragraph{Group Sequence Policy Optimization (GSPO).}
GSPO addresses the mismatch between sequence-level rewards and token-level ratios
by defining a \emph{sequence-level} importance ratio via the geometric mean:
\begin{equation}
s_i(\theta)\triangleq
\exp\!\left(
\frac{1}{|y_i|}\sum_{t=1}^{|y_i|}
\log w_{i,t}(\theta)
\right).
\label{eq:gspo_seq_ratio}
\end{equation}
It then applies clipping at the sequence level:
\begin{equation}
\resizebox{0.43\textwidth}{!}{%
    $
\ell^{\textsc{gspo}}_{i}(\theta)\triangleq
\min\!\Big(
s_i(\theta)\hat{A}_i,\;
\mathrm{clip}\!\big(s_i(\theta),\,1-\epsilon,\,1+\epsilon\big)\hat{A}_i
\Big),
$
}
\label{eq:gspo_surrogate}
\end{equation}
with the full expectation/averaging form given in Appendix~\ref{app:baseline_objectives}.

\paragraph{Geometric-Mean Policy Optimization (GMPO).}
GMPO also leverages geometric aggregation to reduce sensitivity to extreme token-wise
ratios, but (unlike GSPO) keeps the token-level structure and typically performs
token-wise clipping \emph{before} geometric aggregation. We provide the complete
formulation in Appendix~\ref{app:baseline_objectives}.

%% file: sections_new/instability.tex
\section{Diagnosing Instability in MoE Off-Policy RL}
\label{sec:diagnosis}

In this section, we characterize a failure mode frequently observed when applying
off-policy RL (e.g., GRPO) to MoE language models: training becomes unstable and may
collapse.
Our goal is to provide measurable evidence linking \emph{routing instability}
(router drift between $\theta$ and $\theta_{\text{old}}$) to increasingly \emph{volatile}
off-policy mismatch signals and more frequent activation of clipping mechanisms,
which together contribute to optimization instability and eventual collapse.

\subsection{Symptom: Training Instability and Reward Collapse}
\label{sec:diagnosis_symptom}

We start by illustrating the instability phenomenon on Qwen2.5 MoE trained
with GRPO under the countdown task and rule-based reward protocol.
Following GRPO, for each query $x$ we sample $G$ candidate responses
$\{y_i\}_{i=1}^{G}\sim \pi_{\theta_{\text{old}}}(\cdot\mid x)$ and optimize the objective in Eq. (2) with ratio defined in Eq. (1).%

We operationally define \emph{collapse} as a sharp and sustained drop in validation score/reward accompanied by
abnormally large KL / gradient norms.
As shown in Fig.~\ref{fig:diagnosis_collapse}, GRPO can exhibit abrupt collapse on MoE models.
GSPO mitigates collapse but often remains oscillatory, while GMPO may delay collapse yet can still fail
in long runs.


This section establishes that instability is not an anecdotal artifact: it is a reproducible symptom in MoE off-policy RL and motivates a deeper diagnosis of its underlying cause.

\subsection{Measuring Router Drift via a Router-Shift Ratio}
\label{sec:diagnosis_router_drift}

We next introduce a lightweight statistic to quantify routing instability between the
current policy $\theta=(\phi,\psi)$ and the old policy $\theta_{\text{old}}=(\phi_{\text{old}},\psi_{\text{old}})$.
Let $c_{i,t}=(x,o_{i,<t})$ denote the decoding context at token position $t$ of response $o_i$.
At each MoE layer $\ell\in\{1,\dots,L\}$, the router produces a distribution over experts,
denoted by $r^{(\ell)}_{\phi}(e\mid c_{i,t})$.

\paragraph{Old activated experts.}
For each token $(i,t)$ and layer $\ell$, let $\{e^{(\ell,k)}_{i,t}\}_{k=1}^{K}$ be the top-$K$
expert indices selected by the \emph{old} router $\phi_{\text{old}}$ (i.e., the experts activated
when computing the old-policy log-probabilities). We measure how much the current router
changes its probability mass on these old activated experts.

\paragraph{Router-shift ratio.}
We first compute the layer-wise routing deviation
\begin{equation}
\begin{aligned}
d^{(\ell)}_{i,t}
\triangleq
\frac{1}{K}\sum_{k=1}^{K}
\Bigl|
\log r^{(\ell)}_{\phi}\!\left(e^{(\ell,k)}_{i,t}\mid c_{i,t}\right)
-
\log r^{(\ell)}_{\phi_{\text{old}}}\!\left(e^{(\ell,k)}_{i,t}\mid c_{i,t}\right)
\Bigr|.
\end{aligned}
\label{eq:drift_layer}
\end{equation}
and aggregate it across layers:
\begin{equation}
\Delta_{i,t}
\triangleq
\frac{1}{L}\sum_{\ell=1}^{L} d^{(\ell)}_{i,t}.
\label{eq:drift_token}
\end{equation}
We then define the \emph{router-shift ratio} as a bounded coefficient
\begin{equation}
\gamma_{i,t}
\triangleq
\exp(-\Delta_{i,t}) \in (0,1],
\label{eq:router_shift_ratio}
\end{equation}
where larger routing deviations yield smaller $\gamma_{i,t}$.

\paragraph{Logged severity statistics.}
In our large-scale GRPO runs, we log router-shift statistics as detached diagnostics.
For numerical stability, we apply a floor $\gamma_{\min}$ (we use $\gamma_{\min}=0.8$ in our implementation):
\begin{equation}
\bar\gamma_{i,t}\triangleq \max(\gamma_{i,t},\gamma_{\min}),
\mathrm{ClipFrac}_{\gamma_{\min}}\triangleq \Pr(\gamma_{i,t}<\gamma_{\min}).
\label{eq:router_shift_floor}
\end{equation}
Intuitively, $\mathrm{ClipFrac}_{\gamma_{\min}}$ measures the fraction of tokens whose routing deviation
is severe enough to fall below the threshold $\gamma_{\min}$.
We use these statistics to track how routing instability evolves during training.

\subsection{Router Drift Amplifies Off-Policy Mismatch and Triggers Clipping Instability}
\label{sec:diagnosis_mismatch}

As shown in Fig.~\ref{fig:Qwen3_training_rewards}, GRPO training on Qwen3-30B-A3B for math reasoning
exhibits a clear reward-collapse behavior at scale.
We next analyze training-time stability signals to characterize how routing instability relates to
off-policy optimization dynamics.

\paragraph{Routing-side instability.}
The top row of Fig.~\ref{fig:mech_diag} tracks routing-severity signals derived from the router-shift ratio.
Under GRPO, the router-shift ratio decreases while the router-shift clip fraction increases over training,
indicating that routing deviations across policy updates become progressively more severe.

\paragraph{Optimization-side instability.}
The bottom row of Fig.~\ref{fig:mech_diag} reports two lightweight optimization diagnostics:
the importance-ratio signal (logged as \texttt{ppo\_kl}) and the clipping activity \texttt{pg\_clipfrac}.
As routing drift accumulates, the importance-ratio signal becomes increasingly volatile and exhibits
pronounced spikes, accompanied by more bursty clipping.

Together, these patterns suggest an instability cascade in which router drift amplifies off-policy mismatch
and triggers frequent clipping, increasing the risk of training collapse.


\paragraph{Summary and Design Implications.}
\label{sec:diagnosis_summary}

We summarize our diagnosis as follows.
First, MoE off-policy RL exhibits reproducible training instability and reward collapse
(Fig.~\ref{fig:diagnosis_collapse}).
Second, routing stability degrades over training, as reflected by a decreasing router-shift ratio
and an increasing router-shift clip fraction (top row of Fig.~\ref{fig:mech_diag}).
Third, this routing instability coincides with increasingly volatile off-policy mismatch signals
and more frequent activation of clipping constraints (bottom row of Fig.~\ref{fig:mech_diag}),
which together increase the risk of unstable optimization and eventual collapse.

These observations suggest that stabilizing MoE off-policy RL requires directly controlling the impact
of routing drift on off-policy updates while preserving router adaptivity (i.e., without hard-freezing
the router or fully replaying routing decisions). Motivated by this, in the next section we introduce a
router-aware \emph{soft} adjustment that uses the per-token router-shift ratio to down-weight the
importance-ratio contribution of tokens with severe routing deviations.

%% file: sections_new/method.tex
\section{Method}
\label{sec:method}

\subsection{Overview}
\label{sec:method_overview}

Sec.~\ref{sec:diagnosis} shows that, in MoE off-policy RL, routing stability can degrade
across policy updates and is accompanied by increasingly volatile off-policy mismatch signals
and frequent clipping activations, which may culminate in reward collapse.
Existing GRPO-style objectives (e.g., GSPO/GMPO) improve stability mainly through alternative
ratio aggregation and clipping strategies, but they do not explicitly control the impact of
\emph{router drift} on the importance ratio.

We propose \oursfull{} (\ours{}), a lightweight \emph{router-aware} modification that can be
plugged into GRPO and its variants. The key idea is to reuse the per-token router-shift ratio
$\gamma_{i,t}$ defined in Sec.~\ref{sec:diagnosis_router_drift} as a trust signal, and
multiply a processed version of it into the importance ratio before the base algorithm
applies its clipping/aggregation steps.

\subsection{Router-Shift Weight as a Plug-in Rescaling}
\label{sec:method_rsw}

\paragraph{Processed router-shift weight.}
Let $\gamma_{i,t}\in(0,1]$ denote the router-shift ratio defined in
Sec.~\ref{sec:diagnosis_router_drift}.
We apply two practical operations: (i) stop-gradient so it acts purely as a sample weight,
and (ii) flooring to avoid vanishing contributions. Concretely,
\begin{equation}
\tilde{\gamma}_{i,t}
\triangleq
\mathrm{sg}~\!\Big[\max(\gamma_{i,t},\,\gamma_{\min})\Big],
\label{eq:gamma_processed}
\end{equation}
where $\gamma_{\min}\in(0,1]$ is a hyperparameter and $\mathrm{sg}~[\cdot]$ denotes stop-gradient.

\paragraph{Rescaling the importance ratio (before clipping).}
For any GRPO-style objective, define the token-level importance ratio
\begin{equation}
w_{i,t}(\theta)
=
\frac{\pi_{\theta}(o_{i,t}\mid c_{i,t})}
{\pi_{\theta_{\text{old}}}(o_{i,t}\mid c_{i,t})},
~ c_{i,t}=(x,o_{i,<t}).
\label{eq:token_ratio}
\end{equation}
\ours{} replaces $w_{i,t}(\theta)$ with a router-aware adjusted ratio
\begin{equation}
\tilde{w}_{i,t}(\theta)
\triangleq
w_{i,t}(\theta)\cdot \tilde{\gamma}_{i,t},
\label{eq:adjusted_ratio}
\end{equation}
and feeds $\tilde{w}_{i,t}(\theta)$ into the \emph{same} clipping/aggregation pipeline of the
underlying base objective (GRPO/GSPO/GMPO). In other words, \ours{} inserts a single rescaling
step \emph{right before} the base algorithm's clipping, leaving the rest unchanged.

\paragraph{Implementation note (log-space).}
Since ratios are computed in log space in our implementation, Eq.~\eqref{eq:adjusted_ratio}
is implemented stably as
$\log\tilde{w}_{i,t} \leftarrow (\log\pi_{\theta}-\log\pi_{\theta_{\text{old}}}) + \log\tilde{\gamma}_{i,t}$
before exponentiation.

\subsection{Instantiation Used in This Paper}
\label{sec:method_instantiation}

In our main experiments, we instantiate \ours{} on top of GMPO (denoted as \textbf{GMPO+RS},
or \textbf{RSPO}) since GMPO provides a strong and stable GRPO-style base objective for RLVR.
In Sec.~\ref{sec:exp_ablation} we further show that the same router-shift rescaling can also
be plugged into GRPO and GSPO, consistently improving training stability.

%% file: sections_new/exp_new.tex
\section{Experiments}

\subsection{Experimental Setup}
\label{sec:exp_setup}

\paragraph{Models and training regimes.}
We evaluate our method under two complementary regimes.
\textbf{Small-scale diagnostic setting.}
We conduct exploratory experiments and ablations on a Qwen2.5-MoE model pretrained on the Countdown task,
primarily to stress-test training stability and isolate the effect of router-shift weighting.
\textbf{Large-scale benchmark setting.}
For final evaluation, we train Qwen3-30B-A3B on \textbf{math} and \textbf{code} tasks to assess downstream
generalization at scale.

\paragraph{Baselines and hyperparameters.}
We compare against GRPO and two representative GRPO-style variants designed to improve stability: GSPO and GMPO.
For GRPO we adopt the commonly used clipping range $\epsilon{=}0.2$; GSPO/GMPO follow the recommended settings
reported in their respective papers.
All methods are trained under the same rollout budget (8 samples/step) to ensure fair comparison.
For our method, we use a fixed router-shift floor $\gamma_{\min}=0.8$ across both small and large settings and
apply stop-gradient through the router-shift weight when used for optimization.
We report mean results over 3 random seeds for training curves.

\paragraph{Training data and rule-based rewards.}
All settings use verifiable, rule-based rewards (RLVR).
For the small-scale Countdown setting, training data is generated
following the procedure of \cite{qin2025backtrack}.
For large-scale \textbf{math} training, we use DeepScaleR~\cite{luo2025deepscaler}. For large-scale \textbf{code} training, we combine multiple verifiable sources, including PrimeIntellect,
LeetCode, TACO, and LiveCodeBench.

\paragraph{Evaluation protocol.}
For the small-scale setting, we monitor training progress by periodically evaluating on a held-out Countdown test set.
For the large-scale setting, we evaluate both \textbf{math} and \textbf{code}.
For math, we follow the Dr.GRPO protocol and report Pass@1 accuracy on five benchmarks: AIME24,
AMC23, MATH500~\cite{hendrycks2021measuring}, Minerva~\cite{lewkowycz2022solving}, and OlympiadBench~\cite{huang2024olympicarena}; AIME24
results are averaged over 32 runs.
For code, we report Pass@1 on three benchmarks: MBPP, HumanEval, and LiveCodeBench.
Unless otherwise stated, decoding is deterministic (temperature $=0.0$) with one sample per input.

Additional details are provided in Appendix~\ref{app:exp_details}.

\subsection{Main Results}
\label{sec:exp_main}

\begin{table*}[t]
\centering
\caption{\textbf{Main results on Qwen3-30B-A3B.} Pass@1 (\%) on math and code benchmarks.}
\label{tab:main_results}
\resizebox{\linewidth}{!}{
\begin{tabular}{lcccccc|cccc}
\toprule
& \multicolumn{6}{c|}{\textbf{Math}} & \multicolumn{4}{c}{\textbf{Code}} \\
\cmidrule(lr){2-7}\cmidrule(lr){8-11}
\textbf{Method} &
\textbf{AIME24} & \textbf{AMC23} & \textbf{MATH500} & \textbf{Minerva} & \textbf{OlympiadBench} & \textbf{Avg.} &
\textbf{LCB} & \textbf{MBPP} & \textbf{HumanEval} & \textbf{Avg.} \\
\midrule
Base &
80.4 & 90.0 & 90.7 & 47.7 & 62.0 & 74.2 &
52.9 & 86.4 & 83.5 & 74.3 \\
GRPO~\citep{shao2024deepseekmath} &
77.0 & 82.5 & 91.8 & 48.2 & 58.1 & 71.5 &
41.2 & 81.4 & 89.6 & 70.7 \\
GSPO~\citep{zheng2025group} &
80.4 & 95.0 & 93.6 & 48.9 & 64.0 & 76.4 &
58.8 & 87.2 & 95.1 & 80.4 \\
GMPO~\citep{zhao2025geometric} &
80.1 & 92.5 & 94.2 & 49.3 & 65.9 & 76.4 &
64.7 & 87.2 & 95.7 & 82.5 \\
\rowcolor{blue!10}
GMPO+RS (RSPO) &
80.1 & 95.0 & 94.2 & 50.7 & 65.8 & \textbf{77.1} &
70.5 & 88.2 & 97.0 & \textbf{85.2} \\
\bottomrule
\end{tabular}
}
\end{table*}

\begin{figure}[t]
  \centering
  \includegraphics[width=0.7\linewidth]{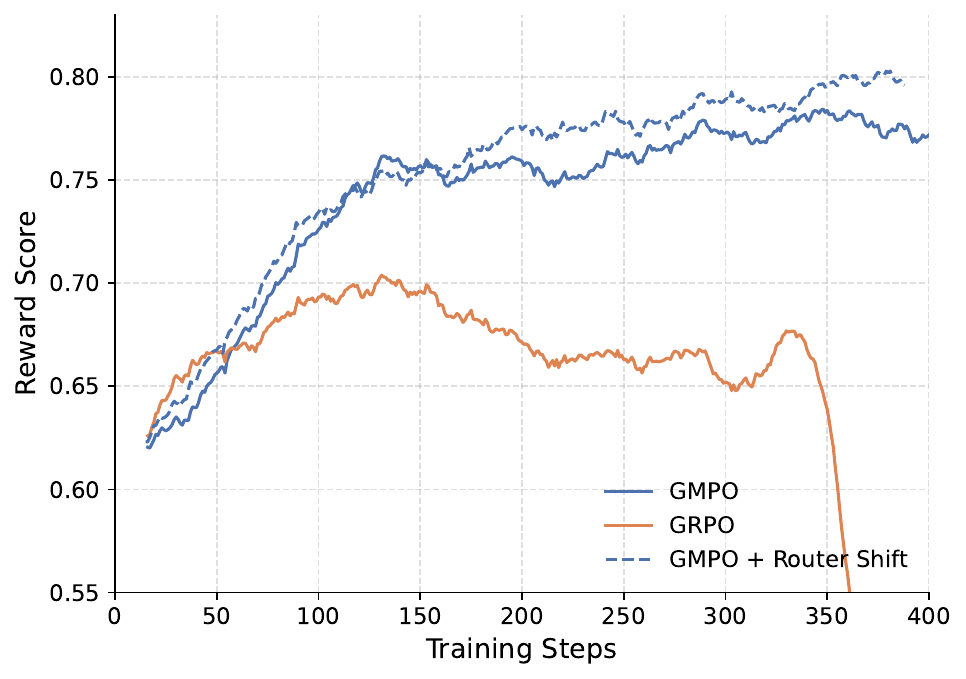}
    \caption{
    \textbf{Training reward dynamics on Qwen3-30B-A3B.}
    Training reward versus training step on the \textbf{math} RLVR setting.
    GRPO exhibits a clear reward collapse in the later stage of training, whereas GMPO remains more stable.
    RSPO (GMPO+RS) maintains stable training and achieves consistently higher reward.
    }
  \label{fig:Qwen3_training_rewards}
\end{figure}

\begin{wrapfigure}{r}{0.5\linewidth}
  \centering
  \vspace{-6pt} 
  \includegraphics[width=\linewidth]{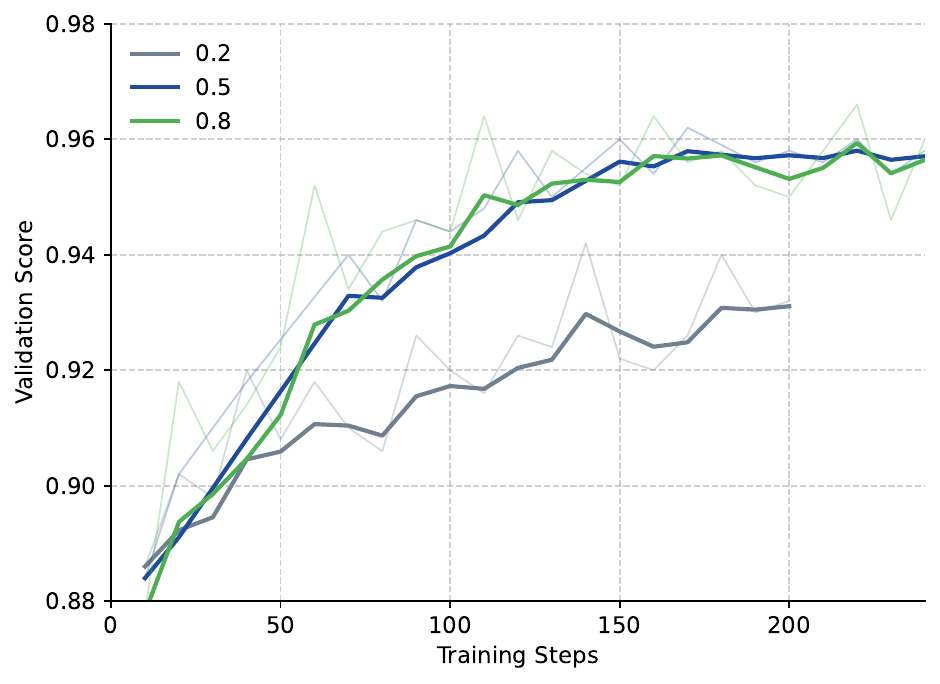}
  \caption{
  \textbf{Sensitivity to the router-shift floor $\gamma_{\min}$ (validation score).}
  Validation score versus training step on Qwen3-30B-A3B under the math RLVR setting.
  Curves correspond to $\gamma_{\min}\in\{0.2,0.5,0.8\}$, with all other hyperparameters fixed.
  }
  \label{fig:gamma_min_sweep}
  \vspace{-6pt}
\end{wrapfigure}

Table~\ref{tab:main_results} summarizes the final Pass@1 performance on Qwen3-30B-A3B after RL training.
On \textbf{math} reasoning, GMPO+RS (RSPO) achieves the best average accuracy (77.1), improving over GMPO/GSPO
(76.4) and GRPO (71.5). The gains are most apparent on challenging benchmarks such as Minerva and
OlympiadBench, while remaining competitive on MATH500 where GMPO/GSPO are already strong.
On \textbf{code}, RSPO yields consistent improvements across all three benchmarks, increasing the average
from 82.5 (GMPO) to 85.2 and substantially outperforming GRPO (70.7). All reported results are averaged over three random seeds.

Figure~\ref{fig:Qwen3_training_rewards} further illustrates training dynamics on Qwen3-30B-A3B.
GRPO exhibits a clear reward collapse in the later stage of training, whereas GMPO is more stable but
converges to a lower reward level.
In contrast, RSPO (GMPO+RS) maintains stable training and achieves the highest reward trajectory,
supporting our claim that router-aware weighting improves both stability and final performance at scale.





\subsection{Ablations}
\label{sec:exp_ablation}

\paragraph{Component contribution: router-shift weighting on top of GMPO.}

To isolate the effect of router-shift weighting in our final method, we compare GMPO with
RSPO (GMPO+RS) under the same large-scale protocol.
As shown in Table~\ref{tab:main_results}, adding router-shift weighting yields consistent improvements:
on \textbf{math}, the average Pass@1 increases from 76.4 (GMPO) to 77.1; on \textbf{code}, it increases
from 82.5 to 85.2.
In addition to improved final accuracy, RSPO exhibits substantially more stable training dynamics
than GRPO and reaches higher reward than GMPO (Fig.~\ref{fig:Qwen3_training_rewards}),
supporting that router-shift weighting provides a complementary stabilization effect beyond geometric
aggregation alone.

\paragraph{Sensitivity to the floor $\gamma_{\min}$.}

We sweep $\gamma_{\min}\in\{0.2, 0.5, 0.8\}$ on Qwen3-30B-A3B and evaluate training progress using
the \emph{validation score} tracked during RL training.
Fig.~\ref{fig:gamma_min_sweep} shows that $\gamma_{\min}=0.5$ and $0.8$ lead to comparable validation
trajectories, while a too-small floor (e.g., $0.2$) can over-suppress tokens with severe routing drift,
weakening learning signals and degrading convergence.
Unless otherwise stated, we use $\gamma_{\min}=0.8$ as the default in all experiments.

\begin{figure}[t]
  \centering
  \includegraphics[width=0.7\linewidth]{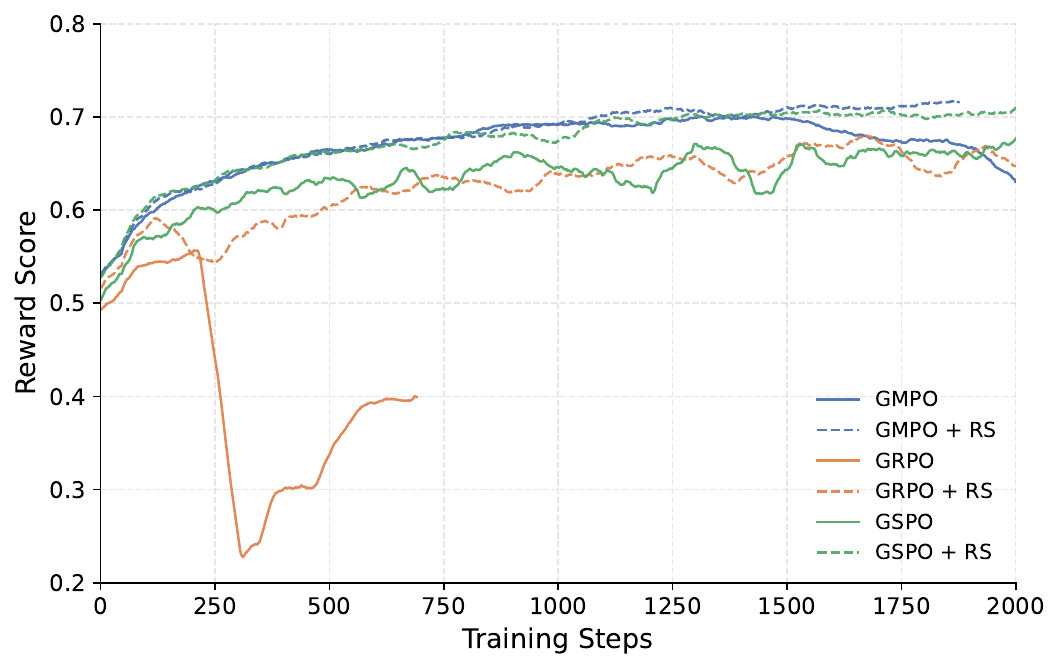}
  \caption{
  \textbf{Router-shift as a plug-in stabilization module (Qwen2.5-MoE, Countdown).}
  Training reward versus step for GRPO/GSPO/GMPO (solid) and their router-shift counterparts (dashed),
  using the same color per base algorithm.
  }
  \label{fig:plugin_small}
\end{figure}

\paragraph{Router-shift as a plug-in component.}
Router-shift weighting is a minimal modification that can be inserted into different GRPO-style objectives.
On the Qwen2.5-MoE Countdown setting, we add the same router-shift weight (Sec.~\ref{sec:method})
to GRPO/GSPO/GMPO while keeping their original clipping/aggregation choices unchanged.
Fig.~\ref{fig:plugin_small} shows that router-shift consistently stabilizes training and improves the
final reward/validation performance, with particularly strong benefits for GRPO which is most prone to
collapse in MoE settings.
This suggests that router-aware weighting is complementary to existing variance-control mechanisms and
can serve as a general stabilization module.

\paragraph{Why stop-gradient on the router-shift weight?}
By default, we treat the router-shift weight as a detached sample weight.
Allowing gradients to flow through the router-shift weight leads to rapid instability in our small-scale
setting; for clarity, we report this ablation in Appendix~\ref{app:stopgrad} (Fig.~\ref{fig:stopgrad_collapse}).
This motivates applying stop-gradient to the router-shift weight throughout the paper.

\subsection{Alternative Router Stabilization Strategies}
\label{sec:exp_alternatives}

We additionally evaluate two intuitive strategies that directly constrain router dynamics:
\textbf{router freezing} and \textbf{routing replay}.
Freezing disables router updates entirely, assuming pretrained routing is already aligned with the RL objective.
Routing replay caches routing decisions from the old policy and reuses them when evaluating the current policy,
thereby eliminating routing drift.

In our experiments, these rigid strategies are not satisfactory:
freezing the router limits adaptation to the RLVR objective, while routing replay restricts router exploration
and incurs non-trivial memory/communication overhead due to caching routing traces across layers and tokens.
In contrast, RSPO achieves stable training without hard constraints by softly down-weighting tokens with severe
routing deviations.
We provide additional comparisons and implementation details for these alternatives in Appendix~\ref{app:alternatives}.


\subsection{Mechanism Diagnostics}
\label{sec:exp_mechanism}

To validate the mechanism identified in Sec.~\ref{sec:diagnosis}, we log lightweight training-time diagnostics
that are available without recording full token-level ratio distributions.
Specifically, we track (i) routing-severity signals from the router-shift ratio (Sec.~\ref{sec:diagnosis_router_drift}),
and (ii) optimization-side signals including the importance-ratio diagnostic (logged as \texttt{ppo\_kl} in our code)
and the PPO/GRPO clipping fraction \texttt{pg\_clipfrac}.
Across runs, we use the same threshold as in training, $\gamma_{\min}=0.8$, when reporting router-shift clip fraction.

\begin{figure*}[t]
  \centering
  \includegraphics[width=\linewidth]{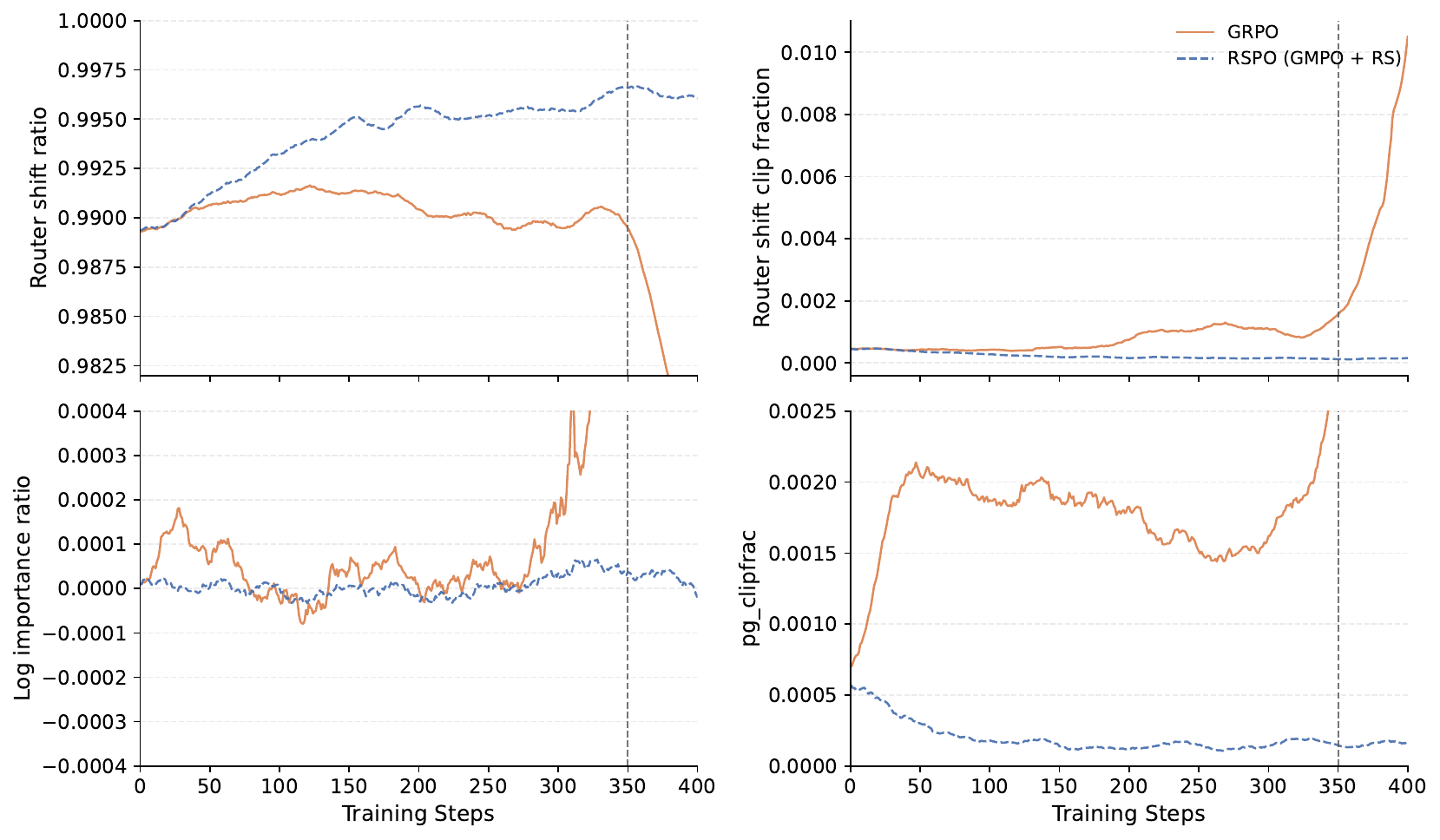}
  \caption{
  \textbf{Mechanism diagnostics on Qwen3-30B-A3B (math RLVR).}
  Router-side severity signals (top row) and optimization-side signals (bottom row) over training for GRPO vs RSPO.
  RSPO keeps routing deviations smaller and reduces volatility and bursty clipping in off-policy updates.
  }
  \label{fig:mech_diag}
\end{figure*}

\paragraph{RSPO stabilizes both router-side and optimization-side signals.}
Fig.~\ref{fig:mech_diag} contrasts GRPO and RSPO on Qwen3-30B-A3B.
Under GRPO, routing stability degrades over training (router-shift ratio decreases and router-shift clip fraction rises),
and the importance-ratio signal becomes increasingly volatile with bursty clipping activity.
In contrast, RSPO maintains substantially more stable routing-severity statistics and reduces the volatility of both
the importance-ratio diagnostic and \texttt{pg\_clipfrac}, consistent with our hypothesis that softly down-weighting
tokens with severe routing drift mitigates the instability cascade that can lead to reward collapse.
We additionally report training entropy as an auxiliary indicator of policy collapse in Appendix Fig.~\ref{fig:entropy}.

\subsection{Efficiency and Overhead}
\label{sec:exp_cost}

RSPO introduces additional overhead mainly from caching routing statistics needed to compute the router-shift
weight (Sec.~\ref{sec:diagnosis_router_drift}) and applying a per-token rescaling to the importance ratio.
On Qwen3-30B-A3B, RSPO incurs a 20.8\% reduction in training throughput
compared to GMPO under the same training configuration,
retaining 79.2\% of GMPO throughput.

In terms of memory, RSPO caches old top-$K$ routing information for each token and MoE layer.
For Qwen3-30B-A3B with $L{=}48$, $K{=}8$, batch size 128, and response length 8192 (about $1.05$M tokens per step),
storing old top-$K$ routing probabilities in FP16 requires approximately 0.75\,GiB per device.
Storing the corresponding expert indices requires an additional 0.75\,GiB if stored as 16-bit integers
(since there are 128 experts), yielding about 1.5\,GiB extra memory in total.
This overhead scales linearly with the per-device batch size and sequence length, and can be reduced by using
compact index types and/or offloading cached indices to CPU memory.

%% file: sections/related.tex
\section{Related Work}
\subsection{Reinforcement learning for LLM}
Recently, the emergence of DeepSeek R1~\citep{guo2025deepseek} has demonstrated the significant potential of combining reinforcement learning (RL) with reasoning for pushing the performance boundaries of large language models (LLMs). At the core of R1 lies the Group Relative Policy Optimization (GRPO)~\citep{shao2024deepseekmath} algorithm, which represents an improvement over the well-known Proximal Policy Optimization (PPO)~\citep{schulman2017proximal} algorithm. GRPO estimates advantages within groups, thereby eliminating the need for an expensive value function model while maintaining performance comparable to PPO.The success of R1 has sparked widespread interest in GRPO and inspired the development of numerous variants. For instance, DAPO~\citep{yu2025dapo} introduces techniques such as dynamic sampling and higher clipping thresholds, addressing challenges related to training efficiency and stability. Dr.\,GRPO~\citep{liu2025understanding} focuses on mitigating length bias by removing the length and standard deviation normalization terms in GRPO, thereby reducing optimization bias and improving token efficiency.More recently, several studies have highlighted issues with the token-level importance sampling ratio used in GRPO, which can lead to increased variance. To address this, GMPO~\citep{zhao2025geometric} proposes maximizing token-level rewards using a geometric mean, resulting in more stable training dynamics. Similarly, GSPO~\citep{zheng2025group} approaches the problem from the sequence-level importance ratio perspective, ultimately also converging on a geometric mean formulation for enhanced stability. Notably, GSPO reports that this geometric mean approach is particularly effective for reinforcement learning training in Mixture-of-Experts (MoE) models.

\subsection{Stability in MoE Training}
Mixture-of-Experts (MoE) models have emerged as a key technique for scaling neural networks to trillions of parameters while maintaining computational efficiency by sparsely activating only a small subset of experts per token. 
However, this sparse activation introduces unique challenges, including expert under-utilization, load imbalance, and routing instability. 
Severe load imbalance can lead to some experts being overloaded while others receive few or no tokens, resulting in inefficient use of model capacity and degraded convergence. 
Switch Transformer~\citep{fedus2022switch} addresses these challenges by introducing an auxiliary load-balancing loss to encourage uniform expert utilization and a capacity factor to cap the number of tokens routed to each expert, thus preventing overload. 
While effective, large auxiliary losses can introduce non-negligible gradient interference with the main training objective. 
\citep{wang2024auxiliary} mitigate this by proposing the Loss-Free Balancing method, which dynamically adjusts expert-wise biases on routing scores before top-$k$ selection to balance expert loads without introducing additional loss terms, thereby avoiding gradient interference and improving the attainable model performance. 
StableMoE~\citep{dai2022stablemoe} further identifies routing fluctuation as a key source of instability, proposing to distill a stable teacher router and freeze it during training to reduce token assignment variance. 
Another line of work focuses on improving gradient flow through non-differentiable top-$k$ routing by using differentiable relaxations such as Gumbel-Softmax or straight-through estimators~\citep{wang2024remoe,puigcerver2023sparse,zhou2022mixture}, thereby reducing gradient variance and enabling end-to-end optimization. 

More recently, researchers have observed that MoE models are particularly unstable under reinforcement learning (RL) training, where reward sparsity and high-variance policy gradients exacerbate routing fluctuations. 
To address this, several approaches aim to stabilize MoE routers during RL fine-tuning. 
For instance, GSPO~\citep{zheng2025group} stabilizes off-policy updates by reusing expert assignments from previous policies and clipping sequence-level importance sampling ratios, effectively reducing update variance. 
Ring-lite~\citep{team2025ring} introduces constrained token-level routing budgets to regularize expert selection and further reduce variance. 
Despite these advances, understanding the interplay between routing dynamics, gradient variance, and RL credit assignment remains an open research direction, motivating methods like RSPO that explicitly account for router shift when shaping policy updates.

%% file: sections_new/conclu.tex
\section{Conclusion}

We studied a practical instability in off-policy RL training for Mixture-of-Experts language models.
Our diagnosis indicates that router drift across policy updates co-occurs with volatile off-policy mismatch signals and bursty clipping activity, which can culminate in reward collapse.
Motivated by this mechanism, we proposed \textbf{Router-Shift Policy Optimization (RSPO)}, a lightweight router-aware modification to GRPO-style objectives that computes a per-token router-shift ratio from old activated experts, applies stop-gradient and a lower-bound floor, and rescales importance ratios before clipping/aggregation.
This soft adjustment preserves router adaptivity while reducing the impact of tokens with severe routing deviations.

Empirically, router-shift weighting acts as a plug-in stabilizer on Qwen2.5-MoE (Countdown), improving stability for GRPO/GSPO/GMPO, and RSPO (GMPO+RS) yields consistent gains on Qwen3-30B-A3B across both math and code benchmarks.
More broadly, our results suggest that explicitly accounting for router dynamics is a key design principle for stable MoE post-training.

%% file: sections_new/append.tex
\section{Experimental Details}
\label{app:setup_details}

\subsection{Models and Training Configurations}
\label{app:exp_details}

\paragraph{Small-scale setting (Qwen2.5-MoE, Countdown).}
The Qwen2.5-MoE model used in the small-scale diagnostic setting contains 12 Transformer layers.
Each MoE layer has 8 experts and activates 1 expert per token (top-$k{=}1$).
The model is pretrained on the Countdown task; the Countdown dataset is generated following the procedure
described in \cite{qin2025backtrack}.
For RL training in this setting, the maximum response length is 8K tokens.

\paragraph{Large-scale setting (Qwen3-30B-A3B, Math/Code).}
The Qwen3-30B-A3B model contains 48 Transformer layers ($L{=}48$).
Each MoE layer has 128 experts and activates 8 experts per token ($K{=}8$).
For RL training in this setting, the maximum response length is 8K tokens.

\paragraph{Batch sizes and rollout group size.}
Across both settings, we use rollout group size $G{=}8$.
For the small-scale setting, the global training batch size is 256 with mini-batch size 64.
For the large-scale setting, the global training batch size is 128 with mini-batch size 64.

\subsection{Algorithms and Hyperparameters}
\label{app:algo_hparams}

We use the same algorithmic choices across small and large settings unless otherwise stated.
For GRPO, we use symmetric clipping with $\epsilon_{\text{low}}=\epsilon_{\text{high}}=0.2$.
For GMPO and GSPO, we follow the clipping ranges recommended in their original papers:
GMPO uses $(\epsilon_{\text{low}},\epsilon_{\text{high}})=(e^{-0.4},\,e^{0.4})$, and
GSPO uses $(\epsilon_{\text{low}},\epsilon_{\text{high}})=(3\times 10^{-4},\,4\times 10^{-4})$.
For RSPO, we fix the router-shift floor to $\gamma_{\min}=0.8$ in all experiments and apply stop-gradient
through the router-shift weight when used for optimization (Sec.~\ref{sec:method}).

\subsection{Rule-based Reward Verifiers}
\label{app:verifiers}

All tasks use verifiable, rule-based rewards (RLVR) with binary rewards in $\{0,1\}$.
For Countdown, rewards are computed by directly matching the model output to the target format/answer.
For math reasoning, we verify final answers using a deterministic math verifier (\texttt{math\_verify}).
For code, we execute generated programs in a sandbox environment against unit tests; reward is 1 if all tests pass
(and the program runs successfully), and 0 otherwise.

\subsection{Evaluation Protocol Details}
\label{app:eval_details}

\paragraph{Math benchmarks.}
We follow the Dr.GRPO evaluation protocol and report Pass@1 accuracy.
AIME24 results are averaged over 32 repeated evaluations.
Decoding is deterministic with temperature $=0$.

\paragraph{Code benchmarks.}
We report Pass@1 on MBPP, HumanEval, and LiveCodeBench.
For LiveCodeBench, we evaluate using the v4--v5 benchmark suite.
For training-time LiveCodeBench data, we use problems released prior to v5 to avoid leakage.
Decoding uses temperature $=0$ and maximum generation length of 32K tokens.

\section{Stop-Gradient on the Router-Shift Weight}
\label{app:stopgrad}
\begin{figure}[t]
\centering
\includegraphics[width=0.7\linewidth]{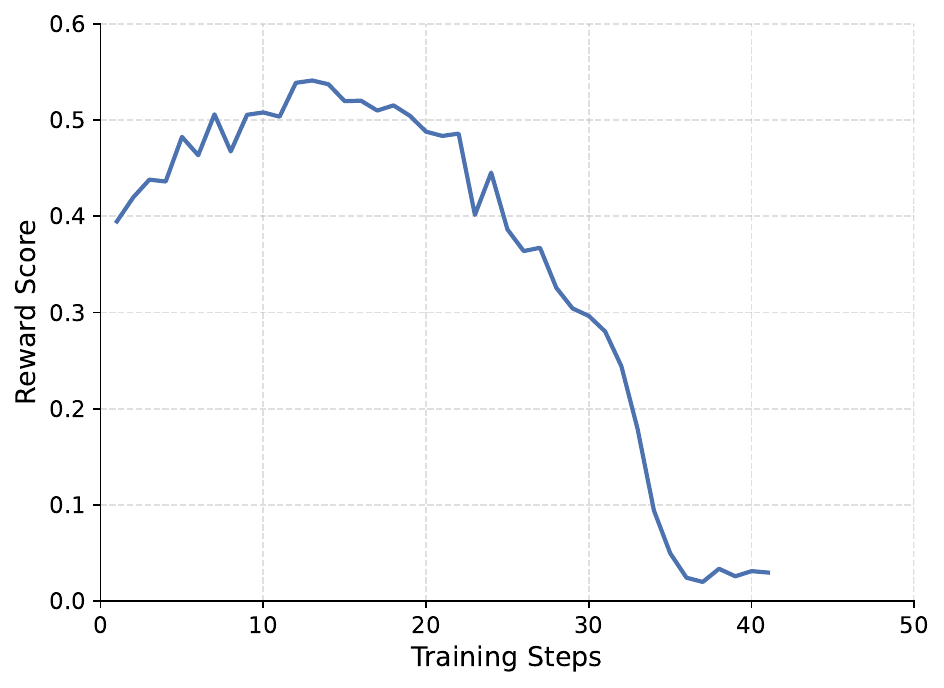}
\caption{
\textbf{Backpropagating through the router-shift weight leads to early collapse.}
Reward/validation score versus step when gradients are allowed to flow through the router-shift weight.
For readability, we plot only the unstable run; in contrast, the default detached setting remains stable
throughout training (see main text).
}
\label{fig:stopgrad_collapse}
\end{figure}

As shown in Figure.~\ref{fig:stopgrad_collapse} on the small Qwen2.5-MoE, backpropagating through \(\gamma\) triggers \textbf{early collapse} in both reward and validation curves, whereas the \textbf{stop-grad} setting yields smooth and stable optimization.
Intuitively, since \(\gamma=\exp(-\lvert\Delta\log r\rvert)\) aggregates layer-wise routing drift, letting \(\partial\log\gamma/\partial\theta\) flow couples the router-shift penalty with the sequence-level geometric objective and clipping, thereby amplifying variance under non-smooth top-\(K\) routing.

\section{Training Entropy}
\label{app:entropy}

\begin{figure}[t]
  \centering
  \includegraphics[width=0.7\linewidth]{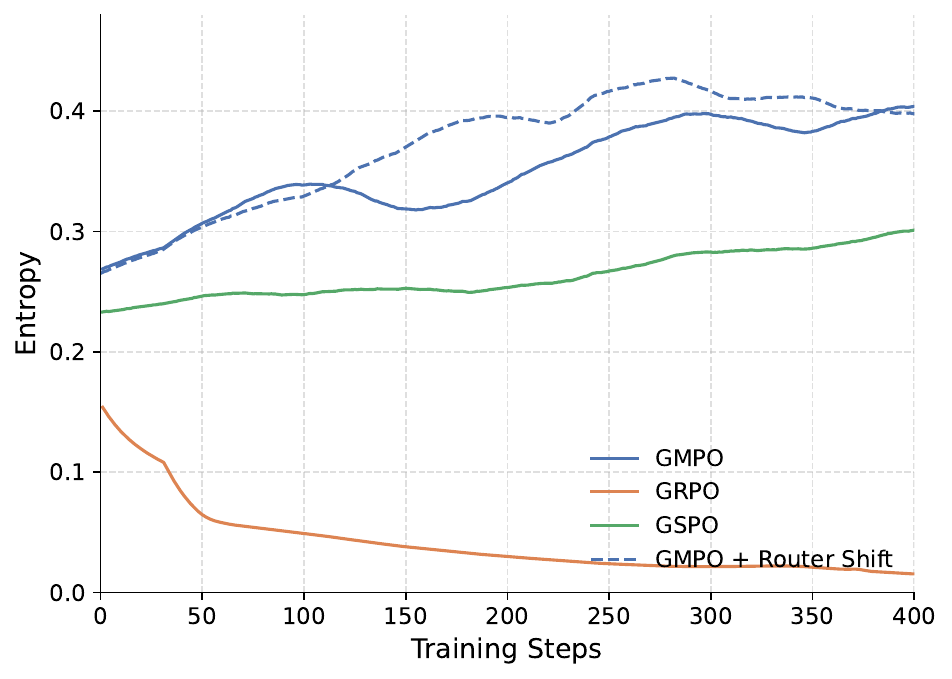}
  \caption{\textbf{Training entropy on Qwen3-30B-A3B (math RLVR).}
  Average token-level policy entropy during training for GRPO, GSPO, GMPO, and RSPO (GMPO+RS).
  GRPO rapidly collapses to near-zero entropy, while the other methods maintain higher entropy.}
  \label{fig:entropy}
\end{figure}

We report training-time policy entropy as an auxiliary indicator of distribution collapse.
A sharp drop in entropy suggests the policy becomes overly deterministic (mode collapse), which often
co-occurs with unstable optimization.
Fig.~\ref{fig:entropy} shows that GRPO quickly exhibits a dramatic entropy decay, whereas GSPO/GMPO and
RSPO maintain substantially higher entropy throughout training, consistent with improved stability.

\section{Additional Attempts on Router Stabilization}
\label{app:alternatives}

In addition to RSPO, we explored several heuristic strategies that aim to stabilize MoE RL training by
\emph{explicitly constraining} router dynamics.
This appendix provides implementation details and empirical observations for these alternatives,
which complement the brief discussion in Sec.~\ref{sec:exp_alternatives}. 

All experiments in this section are conducted in the small-scale diagnostic setting
(Qwen2.5-MoE on Countdown) under the same RLVR protocol as Sec.~\ref{sec:exp_setup}.

\paragraph{(i) Freezing the router.}
A straightforward approach is to freeze the router parameters throughout RL training,
i.e., keeping $\phi$ fixed (no router updates) while updating the remaining parameters.
This removes router drift by construction, but implicitly assumes that the pretrained routing is already
well aligned with the RL objective. In practice, freezing reduces the model's ability to adapt expert
allocation to the evolving policy updates and rewards, which can limit performance.

\paragraph{(ii) Routing replay with logit copying (logit replay).}
Motivated by the routing replay idea discussed in GSPO, we implement a variant that directly reuses the
\emph{old} router logits when evaluating the \emph{current} policy during optimization.
Concretely, during the update step, we replace the current router logits (or routing scores) with those
cached from the old policy $\phi_{\text{old}}$, so that both expert selection and expert weighting are
fully aligned with the old router.
A drawback is that the current router is no longer used to compute routing logits, so gradients cannot
propagate to the router parameters, effectively preventing router learning.

\paragraph{(iii) Routing replay with expert-index reuse (index replay).}
As an alternative, we cache only the old top-$K$ expert indices $\{e^{(\ell,k)}_{i,t}\}_{k=1}^{K}$
selected by $\phi_{\text{old}}$ and enforce the current policy to route to these stored indices during
the update.
Unlike logit replay, the current router still computes its own routing scores on the reused indices,
but the discrete expert choices are constrained.
This preserves the old routing support while allowing limited router gradients, at the cost of restricting
router exploration and potentially introducing mismatch when the optimal routing changes.

\begin{figure}[t]
  \centering
  \includegraphics[width=0.8\linewidth]{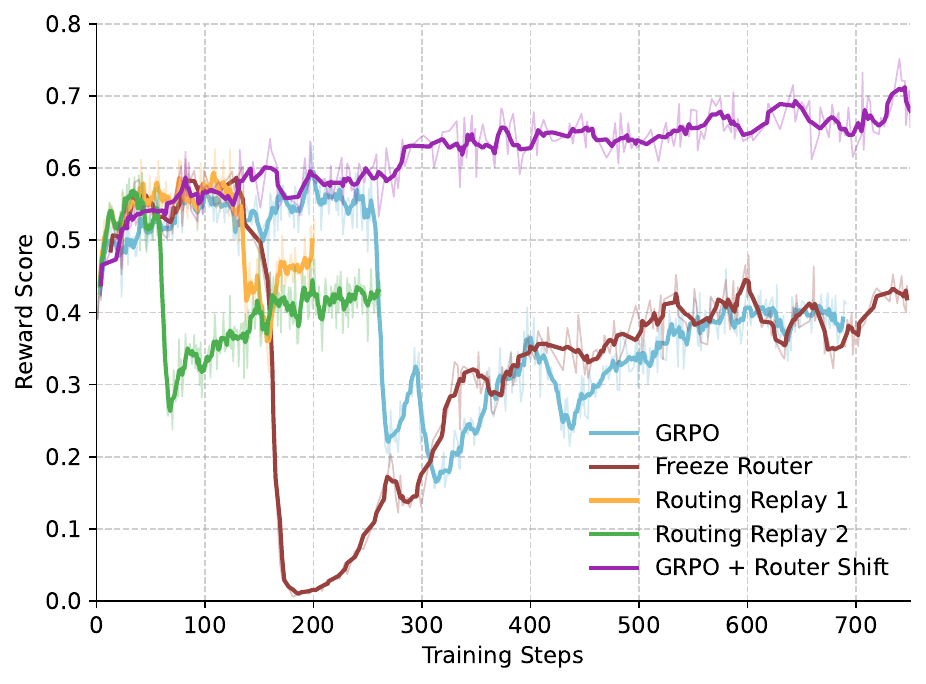}
  \caption{
  \textbf{Alternative router stabilization strategies on Qwen2.5-MoE (Countdown).}
  Training reward (or validation score) versus training step for (i) freezing the router,
  (ii) routing replay by copying old router logits (logit replay), and
  (iii) routing replay by reusing old expert indices (index replay).
  }
  \label{fig:router_replay_freeze}
\end{figure}

\paragraph{Empirical results.}
Fig.~\ref{fig:router_replay_freeze} summarizes the training dynamics under these three strategies
(router freezing, logit replay, and index replay).
Overall, none of the rigid approaches consistently improves stability or final performance compared with
our soft router-aware adjustment:
freezing limits adaptation, while replay-based variants constrain router learning/exploration and may
still exhibit unstable optimization behavior.

\paragraph{Practical considerations.}
Routing replay requires caching routing traces (logits or indices) across tokens and MoE layers, which can
incur non-trivial memory and communication overhead in distributed training.
In contrast, RSPO uses only lightweight statistics on old activated experts and applies a detached
per-token weight, achieving stabilization without hard constraints.

\section{Full Objectives of GRPO-Style Baselines}
\label{app:baseline_objectives}

\paragraph{GRPO.}
Given $x\sim\mathcal{D}$, sample $\{y_i\}_{i=1}^{G}\sim \pi_{\theta_{\text{old}}}(\cdot\mid x)$.
The GRPO objective averages the token-level clipped surrogate:
\begin{equation}
\begin{aligned}
\mathcal{J}_{\textsc{grpo}}(\theta)
=
\mathbb{E}_{x,\{y_i\}}
\Bigg[
\frac{1}{G}\sum_{i=1}^{G}
\frac{1}{|y_i|}\sum_{t=1}^{|y_i|}
\ell^{\textsc{grpo}}_{i,t}(\theta)
\Bigg],
\end{aligned}
\label{eq:grpo_full_obj}
\end{equation}
where $w_{i,t}(\theta)$ is defined in Eq.~\eqref{eq:grpo_token_ratio} and
$\ell^{\textsc{grpo}}_{i,t}(\theta)$ is defined in Eq.~\eqref{eq:grpo_surrogate}.
The group-relative advantage $\hat{A}_i$ is computed by normalizing rewards within
the group (mean/std over $\{r(x,y_i)\}_{i=1}^{G}$).

\paragraph{GSPO.}
GSPO computes the sequence-level ratio $s_i(\theta)$ (Eq.~\eqref{eq:gspo_seq_ratio})
and applies sequence-level clipping:
\begin{equation}
\begin{aligned}
\mathcal{J}_{\textsc{gspo}}(\theta)
=
\mathbb{E}_{x,\{y_i\}}
\Bigg[
\frac{1}{G}\sum_{i=1}^{G}
\ell^{\textsc{gspo}}_{i}(\theta)
\Bigg],
\end{aligned}
\label{eq:gspo_full_obj}
\end{equation}
where $\ell^{\textsc{gspo}}_{i}(\theta)$ is defined in Eq.~\eqref{eq:gspo_surrogate}.

\paragraph{GMPO.}
GMPO uses geometric aggregation to form a robust sequence-level ratio, commonly by
clipping token-wise ratios before aggregating:
\begin{equation}
\begin{aligned}
\bar{s}_i(\theta)
\triangleq
\exp\!\left(
\frac{1}{|y_i|}\sum_{t=1}^{|y_i|}
\log \mathrm{clip}\!\big(w_{i,t}(\theta),\,\epsilon_1,\,\epsilon_2\big)
\right),
\end{aligned}
\label{eq:gmpo_ratio}
\end{equation}
and then optimizes a GRPO-style surrogate using $\bar{s}_i(\theta)$ and $\hat{A}_i$
(the exact clipping bounds follow each method's recommended settings).